\documentclass[letterpaper, 10 pt, conference]{ieeeconf}  

\IEEEoverridecommandlockouts                              

\usepackage{graphicx}

\usepackage{xcolor}
\usepackage{soul}
\usepackage{booktabs}

\newcommand{\norm}[1]{\left\lVert#1\right\rVert}

\usepackage{float}
\usepackage{caption}
\usepackage{subcaption}

\usepackage{multicol}
\usepackage{multirow}

\usepackage{makecell}

\usepackage{mathtools}
\usepackage[export]{adjustbox}
\usepackage{fixltx2e}
\usepackage{hyperref}

\usepackage{amsmath}
\usepackage{amsfonts}
\usepackage{cite}

\title{\LARGE \bf
Explosive Jumping with Rigid and Articulated Soft Quadrupeds via Example Guided Reinforcement Learning
}

\author{Georgios~Apostolides$^{1}$,
        Wei Pan$^{2}$, Jens Kober$^{1}$, Cosimo~Della~Santina$^{1,3}$, and Jiatao~Ding$^{\star,4}$
\thanks{The work is supported in part by EU project INVERSE (GA 101136067).}
\thanks{Georgios~Apostolides, Jens Kober and Cosimo Della Santina are with Cognitive Robotics, Delft University of Technology, Building 34, Mekelweg 2, 2628 CD Delft, Netherlands (e-mail: gapostolides@hotmail.com, J.Kober@tudelft.nl, C.DellaSantina@tudelft.nl).
Wei Pan is with the Department of Computer Science, University of Manchester, UK (email: wei.pan@manchester.ac.uk). Jiatao Ding is with the Department of Industrial Engineering, University of Trento, via Sommarive, 9, 38123 Trento, Italy (jiatao.ding@unitn.it).
Cosimo Della Santina is also with the Institute of Robotics and Mechatronics, German Aerospace Center (DLR), 82234 Wessling, Germany (e-mail: cosimodellasantina@gmail.com).}
\thanks{Jiatao Ding is the corresponding author.}
}

\begin{document}

\maketitle
 \thispagestyle{empty}
\pagestyle{empty}

\begin{abstract}
Achieving controlled jumping behaviour for a quadruped robot is a challenging task, especially when introducing passive compliance in mechanical design.
This study addresses this challenge via imitation-based deep reinforcement learning with a progressive training process. To start, we learn the jumping skill by mimicking a coarse jumping example generated by model-based trajectory optimization. Subsequently, we generalize the learned policy to broader situations, including various distances in both forward and lateral directions, and then pursue robust jumping in unknown ground unevenness. In addition, without tuning the reward much, we learn the jumping policy for a quadruped with parallel elasticity. Results show that using the proposed method, \textit{i)} the robot learns versatile jumps by learning only from a single demonstration,  \textit{ii)} the robot with parallel compliance reduces the landing error by 11.1\%, saves energy cost by 15.2\% and reduces the peak torque by 15.8\%, compared to the rigid robot without parallel elasticity, \textit{iii)} the robot can perform jumps of variable distances with robustness against ground unevenness (maximal $\pm4$cm height perturbations) using only proprioceptive perception. 

\end{abstract}

\section{INTRODUCTION}
\begin{figure*}[t]
    \centering
    \includegraphics[width=0.82\textwidth]{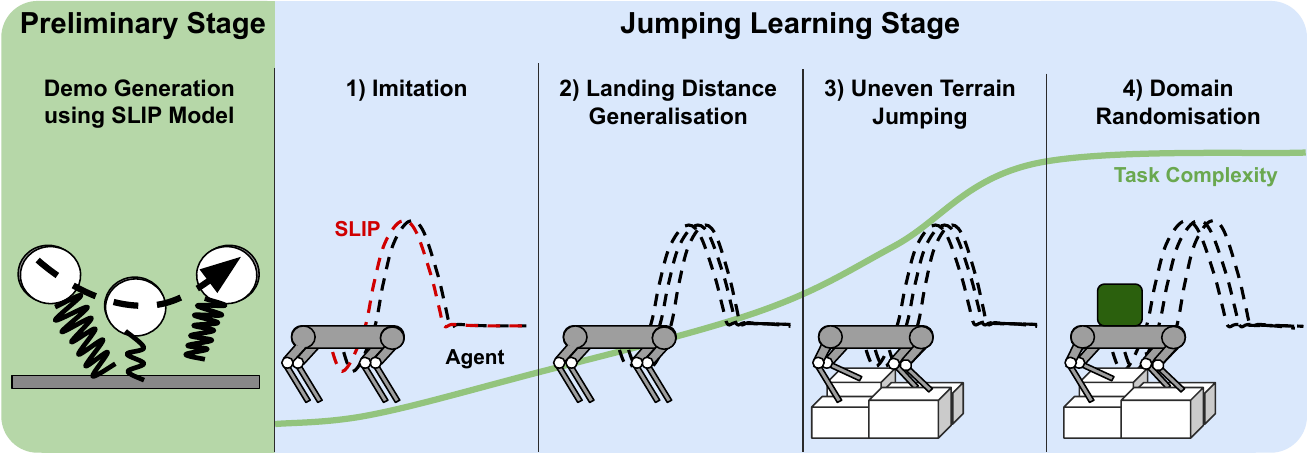}
    \caption{An overview of the progressive learning strategy.}
    \label{fig:methodology-stages}
    \vspace{-5mm}
\end{figure*}

Quadrupedal locomotion has been well investigated in past decades. Introducing passive compliance enhanced the performance \cite{hutter2016anymal,sprowitz2018oncilla,seidel2020using,arm2019spacebok,badri2022birdbot,ruppert2022learning,bjelonic2023learning,ding2024quadrupedal}, among which parallel compliance could improve the energy efficiency and enhance the explosive jumping behavior \cite{arm2019spacebok,badri2022birdbot,ding2024quadrupedal,bjelonic2023learning,ruppert2022learning}.  However, achieving controlled jumping in real environments with uncertainties, especially with the compliant quadruped, is still an open challenge.

To accomplish a successful jumping, model-based approaches have been proposed, where an accurate model is required to capture the dynamics property \cite{dario_bellicoso_dynamic_2017,ernst2010spring}. By using advanced optimization techniques such as trajectory optimization (TO), these strategies are proven effective in achieving stable locomotion \cite{dario_bellicoso_dynamic_2017} or even highly dynamic skills such as jumping \cite{nguyen_optimized_2019,ding_kinodynamic_2020,song2022optimal}. However, this type of method is especially computationally demanding. When it comes to uncertainties, such as uneven terrains in environments, these methods tend to struggle, requiring a dedicated design and tedious tune. Furthermore, when parallel compliance is introduced, the problem becomes more challenging since it is hard to obtain full-body dynamics \cite{ding2024robust}. 

In contrast to model-based approaches, model-free reinforcement learning (RL) has gained particular prominence in recent years. Through learning with interaction, RL-based approach is 
 able to realize quadrupedal locomotion not only on flat ground \cite{peng_learning_2020} but also over uneven terrain \cite{lee_learning_2020}. 
However, to accomplish a jumping task from scratch, a significant hurdle emerges in reward shaping or training design \cite{atanassov2024curriculum,li2023learning,vezzi2024two}. The introduction of expertise reference helps to simplify the reward engineering \cite{peng_deepmimic_2018}. However, in order to enhance generalization, most of the example-guided RL (ERL) methods require multiple demonstrations \cite{fuchioka2023opt,bellegarda_robust_2020,nguyen_continuous_2022}, which may be a difficult task itself. Although the work, such as \cite{peng_deepmimic_2018} and \cite{peng_learning_2020}, only require a single demonstration, they often confine the learning scope to the demonstrated motion. 

Almost all aforementioned ERL-based works assume flat terrain.  An exception work that addresses robust jumping on uneven terrain is \cite{bellegarda2024robust}, where the robot jumps from uneven terrain and lands on a flat surface. However, this work relies on the reference trajectory, which is updated every time a different jumping distance is desired. The work in \cite{li2023learning} showcased jumping onto obstacles of different height variations. However, the robot still jumps and lands on a flat surface. 
To achieve explosive jumping with parallel compliance, the work in \cite{ding_robust_2024} combines TO and 
iterative learning control, realizing robust pronking on uneven ground. However, the work requires to regenerate the reference for each new task, and the optimal utilization of parallel compliance was not investigated. Combing an evolutionary strategy and RL, a two-stage learning framework was proposed in \cite{vezzi2024two}, where, however, the generalization was not reported.


In this work, we introduce a novel ERL approach for explosive jumping. In addition to generalization, our focus also lies on scenarios in which a quadruped robot performs versatile jumps from and onto uneven surfaces, using a single demonstration throughout the entire learning process. To address this challenge, we divide the learning process into five stages: generating a coarse reference in the first preliminary stage, mimicking the reference in the second stage, generalization in the third stage, and robust jumping in the fourth stage. In the final stage, the controller undergoes domain randomization to ensure its robustness in real-world environments. 

The main contributions include
\begin{itemize}
    \item {Learning a jumping policy from a single coarse demonstration:} We learn the complex jumping behaviour from a single demonstration that is generated by trajectory optimization with a reduced-order dynamic model. This single demonstration is used for all tasks in this work. 
    \item {Generalized jumping with robustness in real-world environments:} Through progressive learning, we generalize the learned policy to versatile tasks without revising the reward much. Furthermore, we achieve robust jumping on uneven surfaces.
    \item {Exploiting parallel elastic actuation (PEA):} We train the jumping policy with parallel compliance, achieving explosive jumping with more accurate landing, higher energy cost and lower joint torque.
\end{itemize}

Sect.~\ref{sec:problem_formulation} provides an overview of the methodology 
and Sect.~\ref{sec:method} details the proposed approach. Sect.~\ref{sec:result_discussion} presents the results and Sect.~\ref{sec:conclusion} concludes this work.

\section{Problem formulation} \label{sec:problem_formulation}

\subsection{Preliminary}
\subsubsection{RL basics}
\label{Sec: Notation Overview}
The RL problem is typically formulated as a Markov Decision Process (MDP), where at each step, the agent interacts with the environment by taking an \textbf{action} \(\mathbf{a}_t \in \mathcal{A}\). The environment then transitions to a new \textbf{state} \(\mathbf{s}_{t+1} \in \mathcal{S}\) and provides a corresponding \textbf{reward} \(\mathcal{R}_t\). The agent does not directly observe \(\mathbf{s}_{t+1}\) but instead receives an \textbf{observation} \(\mathbf{o}_{t+1} \in \mathcal{O}\), which is a noisy representation of the true state. Based on \(\mathbf{o}_{t+1}\) and its policy \(\pi(a_{t+1} \mid o_{t+1})\), the agent selects the next action \(\mathbf{a}_{t+1}\). The goal of the RL algorithm is to learn a policy that maximizes the expected cumulative reward over time.

In this work, we adopt the goal-conditioned RL: finding a policy $\pi(a|o,g)$ which maximizes the cumulative sum of rewards earned over the given task $g$. By introducing a discount factor $\gamma \in (0,1]$, our objective is expressed as
\begin{equation}
    \mathop{\arg\max\limits_{\mathbf{\pi}}} \quad J(\pi) = \mathbb{E}_{\tau \sim p^\pi(\tau)}\left[\sum_{t=0}^{T}\gamma^t R_t \vert s_0 = s\right],
\end{equation}
where $R_t$ is the immediate reward at time $t$ and $s_0$ is the initial state. The expectation of the accumulative return $J$ is taken over a trajectory $\tau$ sampled by following the policy $\pi$.

\subsubsection{Reference generation via TO}
\label{Sec: Reference Generation}
\begin{figure}
    \centering
    \includegraphics[width=\columnwidth]{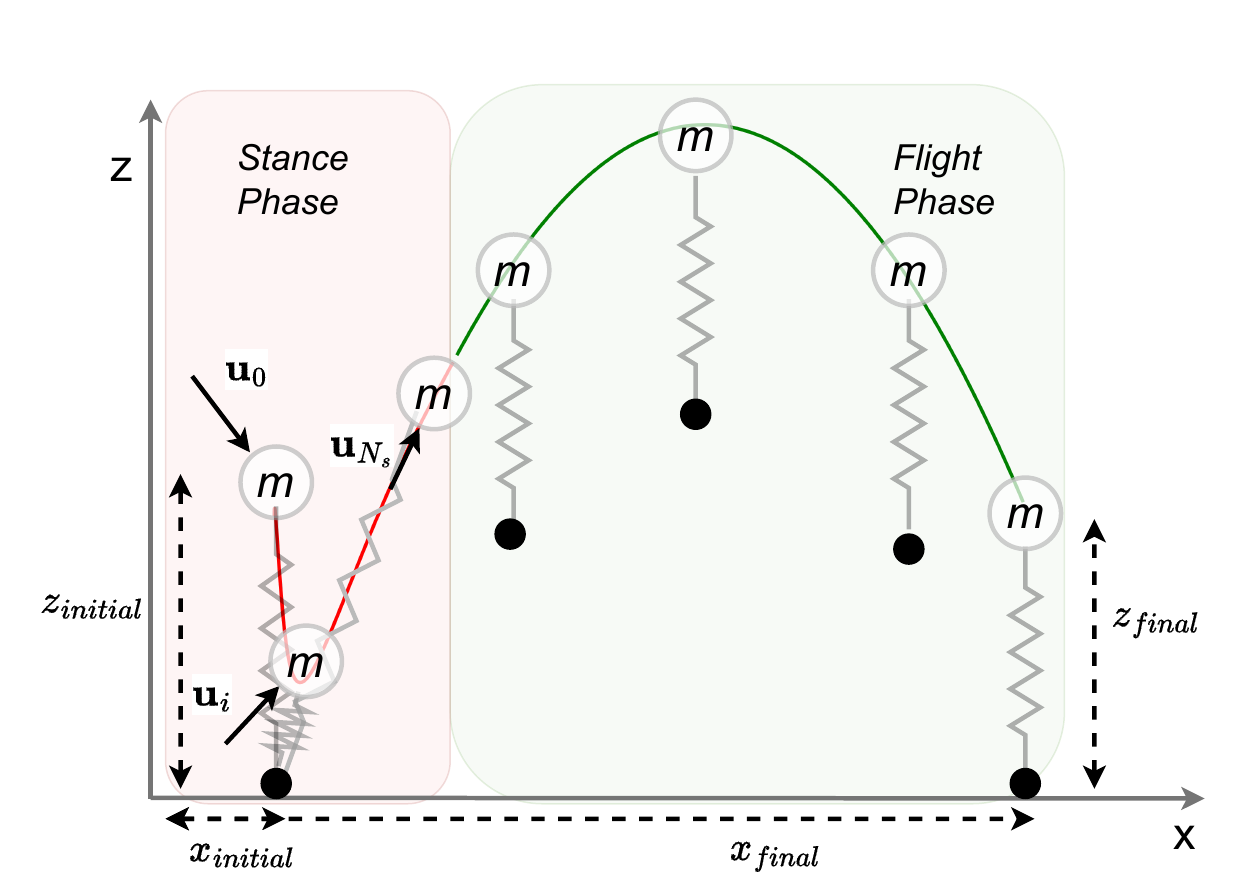}
    \caption{Example trajectory generated via slip-based TO. The sagittal CoM movements (marked by $m$) during the stance and flight phases are shown in red and green, respectively. $\mathbf{u}_i\in {0,\ldots,N_s}$ show the control inputs during stance. The TO problem outputs the CoM position and velocity each discrete step. The foot position (marked by black dots) is only shown for illustration but is otherwise not part of the optimization problem.}
    \label{fig:SLIP}
    \vspace{-8mm}
\end{figure}

In this work, we generate a coarse reference motion via TO. To this end, we use an actuated spring-inverted pendulum (SLIP) model to capture the jumping dynamics \cite{ding2024robust}, where leg mass is ignored. With this model, the body motion is determined by both the spring forces (equal to zeros if no parallel compliance is considered) and actuation forces ($\mathbf{u}_i$ in Fig.~\ref{fig:SLIP})  during the stance phase. In flight, the body motion is totally determined by gravity.

\begin{figure*}[t]
    \centering
    \includegraphics[width=0.93\textwidth]{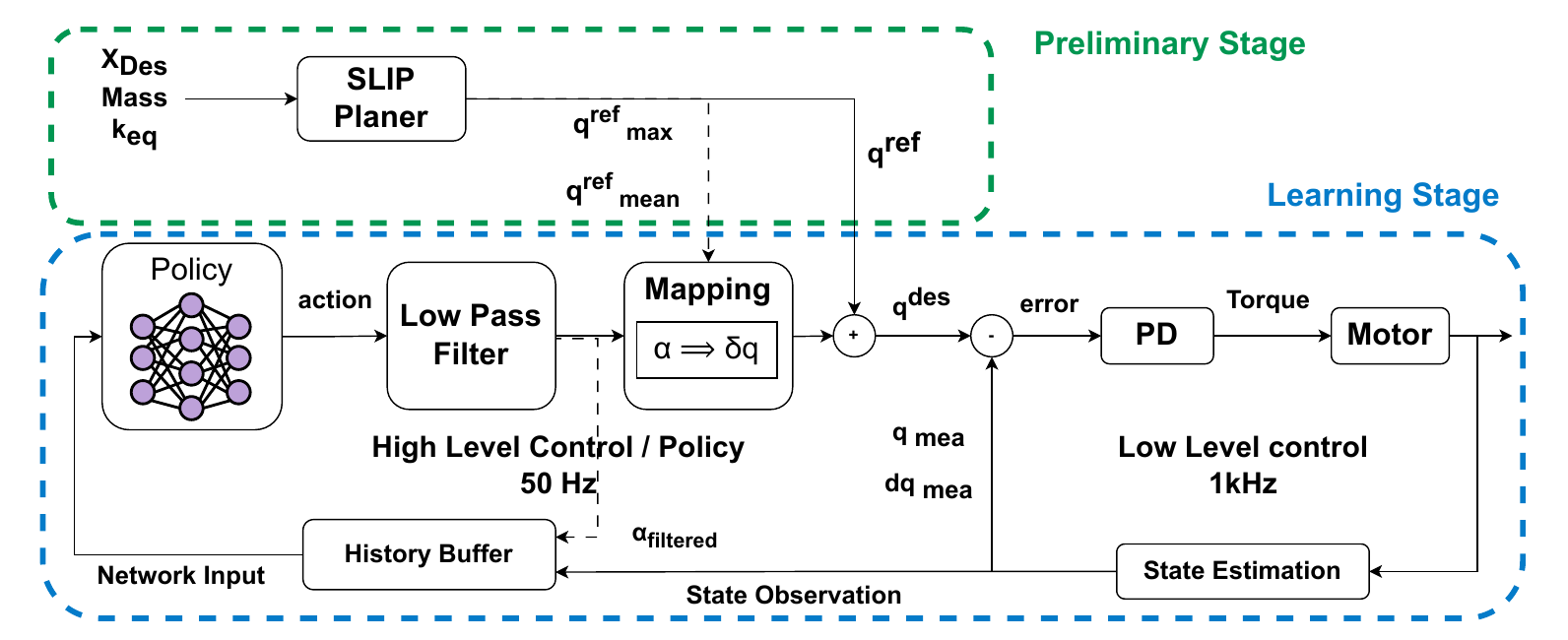}
    \caption{Summarizes the way in which the output of numerical optimization is used by the policy trained using RL, as well as the process which a predicted action undergoes before being converted into a torque that is sent to the robot motors.}
    \label{fig:control-diagram}
    \vspace{-5mm}
\end{figure*}

Given an initial state, we then optimize the jumping motion by defined the reference state, including body position and velocity at key waypoints, such as the state for take-off and landing. Then, a constrained optimization problem is formulated by considering feasibility constraints. The above problem is formulated via the \texttt{CasADi} \cite{andersson2019casadi} library. In addition, the generated trajectory is extended to include a landing phase where the joint angles are fixed. During the entire process, the orientation and the lateral position are set to zeros. Then, the whole jumping trajectory is linearly interpolated to meet the real-time control requirements. Aside from the Cartesian trajectory, the reference joint angles are generated via inverse kinematics. For more details about the TO formulation, please check the work in \cite{ding2024robust}. 


\subsection{Example guided  learning for jumping control}
\label{Sec: Framework Overview}

As visualized in Fig. \ref{fig:methodology-stages}, the entire training process can be decomposed into 4 stages. Before that, the preliminary stage (Sect.~\ref{Sec: Reference Generation}) generates the reference. In the learning process, the first stage imitates the jumping trajectory using ERL (Sec.~\ref{Sec: Imitation Learning}). The second stage generalizes the learned policy to versatile tasks (Sec.~\ref{Sec: Generalisation Stage}). The third stage realizes robust jumping across uneven terrains (Sec.~\ref{Sec: Uneven Terrain Stage}). Finally,  domain randomization makes the jumping more realistic (Sec.~\ref{Sec: Domain randomization}).

As seen in Fig.~\ref{fig:control-diagram}, from the $1^{\text{st}}$ to $4^{\text{th}}$ stage, the policy receives a tuple of observations and actions for the last 20 time-steps as input and outputs a residual action from the reference trajectory.

The \textit{observation} consists of joint angles, joint velocities, Cartesian base velocities, angular velocities and the orientation in quaternion form. In addition, the policy receives a variable (ranges from 0 to 1) that represents the remaining time of the current episode and also a tuple of the desired landing position. The \textit{action} $(\boldsymbol{a_{i}}) \in \mathbb{R}^{12}$ is filtered by a low pass filter and is then converted into a residual joint angle $(\boldsymbol{\delta q}_i)$, which is realized by linearly scaling the filtered action using the max and mean reference joint angles. As a result, the desired joint angle at the $i$-th step $(\boldsymbol{q}^{\text{des}}_i)$ is given by
\begin{equation}
    \boldsymbol{q}^{\text{des}}_i = \boldsymbol{\delta q}_i + \boldsymbol{q}^{\text{ref}}_{i}.
\end{equation}

The desired joint angle is then tracked using a low-level PD controller that is run at 1kHz while the desired joint angle is updated by the policy at 50Hz.

An episode is \textit{terminated early} if there is illegal contact with the ground or if the robot has fallen over. A contact is judged as illegal if any other part of the robot apart from its feet is in contact with the ground. The agent is judged to have fallen if the position vector of the base in the z-direction has an angle bigger than 30 degrees from the global z-direction vector. Additionally, the episode is terminated early during the imitation stage (1$^{\text{st}}$ stage) if the agent has a feet contact state which does not match the contact state of the demonstration for more than 120ms.



\section{Methodology} \label{sec:method}
In this section, we detail the methodology for learning dynamic jumping motion, from the 1$^{\text{st}}$ to 4$^{\text{th}}$ stage. 

\subsection{Stage 1: Imitation stage}
\label{Sec: Imitation Learning}


Following the idea in \cite{peng_deepmimic_2018}, we utilize the reference state initialization (RSI) technique where the agent begins from a randomly sampled state along the desired reference trajectory to start the training. To account for changes in the reward value when the agent starts from a different initial state, we normalize the reward by the number of maximum possible actions that can be taken within the episode length. 

During the imitation stage, the reward is described by
\begin{equation}
    \mathcal{R} = \sum^{N}_{i=0}r^{\text{im}}_i + r^{\text{land}} - p^{\text{smooth}} - p^{\text{land}} + r^{\text{survival}},
\end{equation}

where \( r^{\text{im}} \) is an imitation reward given to the agent at each time step, \( r^{\text{survival}} \) is a survival reward, and \( r^{\text{land}} \) is a reward for landing close to the desired position, both provided given that the agent has survived until the final step. Additionally, \( p^{\text{smooth}} \) and \( p^{\text{land}} \) are penalties applied for smoothing the motion and ensuring stability after landing, respectively.

The imitation reward at the $i$-th step $(r^{\text{im}})$ is calculated  as follows:
\begin{equation}
 r^{\text{im}}_i\!=\!w^{\text{q}}f(k_q, \boldsymbol{\hat{q}}_i,\boldsymbol{q}_i)\!+\! w^{\text{p}}f(k_{p},\boldsymbol{\hat{p}}_i,\boldsymbol{p}_i) \!+\!  w^{\text{o}}f(k_{o}, \boldsymbol{\hat{o}}_i,\boldsymbol{o}_i),
 \label{eq:imitation-reward}
 \end{equation}
where the exponential kernel function is represented by: $f(k,\hat{\boldsymbol{x}}_i,\boldsymbol{x}_i)  = \text{exp}(-k\norm{\boldsymbol{\hat{x}_i}-\boldsymbol{x}_i}_2)$. Here, $\hat{\boldsymbol{x}}$ represents the state of the demonstration while $\boldsymbol{x}$ represents the actual state of the agent. An imitation reward is given for the base position $\boldsymbol{p}_i$, the joint position $\boldsymbol{q}_i$ and the base orientation $\boldsymbol{o}_i$. $w^{\text{q}}$, $w^{\text{p}}$ and $w^{\text{o}}$ are the weightss.  

The landing reward $(r^{\text{land}})$ is given at the end of the episode and is only provided if the episode has not been terminated. The reward is computed by
\begin{equation}
    r^{\text{land}} = w_{\text{land}}f(k_{\text{land}}, \boldsymbol{\hat{p}}_{\textbf{land}},\boldsymbol{p}_\textbf{land}),
    \label{eq:landing-reward}
\end{equation}
where $(\boldsymbol{\hat{p}}_{\textbf{land}})$ and $(\boldsymbol{p_{\textbf{land}}})$ separately denote the desired landing position coming from the demo and the actual landing position. $w_{\text{land}}$ is the weight. The agent is considered to have landed if it passes a flight phase and then at least one leg touches the ground. Once the landing is detected, the base position is recorded as the landed position. 

The smooth penalty ($p^{\text{smooth}}$) is imposed to penalize the joint velocities $(\boldsymbol{\dot{q}})$ and joint accelerations $(\boldsymbol{\ddot{q}})$:
\begin{equation}
    p^{\text{smooth}} = w^{\text{pen}}_{\text{ddq}} \sum_{i=i_{\text{ air}}}^{N}  \norm{\boldsymbol{\ddot{q}}_i}_1 + w^{\text{pen}}_{\text{dq}} \sum_{i=i_{\text{ air}}}^{N} \norm{\boldsymbol{\dot{q}}_i}_1 .
    \label{eq:smoothing-pen}
\end{equation}
with $w^{\text{pen}}_{\text{ddq}}$ and $w^{\text{pen}}_{\text{dq}}$ being the weights.

The stability penalty ($p^{\text{land}}$) is applied on the horizontal velocity $(\boldsymbol{v}^{xy})$ once the robot has landed, computed as
\begin{equation}
    p^{\text{land}}= w^{\text{pen}}_{\text{vel}} \sum_{i=i_{\text{ land}}}^{N} \norm{\boldsymbol{v}^{xy}}_1.
    \label{eq:landing-pen}
\end{equation}
with $w^{\text{pen}}_{\text{vel}}$ being the weight.

Provided that the agent has reached the final step of the episode and an early termination has not been triggered, then the agent will receive a positive survival reward ($r^{\text{survival}}$). In contrast, if there is an early termination, the agent receives a negative reward. Hence, $r^{\text{survival}}$ is tuned by hand,
\begin{equation}
r^{\text{survival}}= 
\begin{cases}
            - 0.1 &   i_\text{ termination} < N\\
            + 0.1 &   i_\text{ termination} = N.
\label{eq:reward-survival}
\end{cases}
\end{equation}

\subsection{Stage 2: Generalization stage}
\label{Sec: Generalisation Stage}
To achieve versatile jumping with different landing positions, the desired distance $(\boldsymbol{\hat{p}}_{\text{land}})$, which is fed as an input to the network, is changed during the generalization stage. The forward desired jumping distance is uniformly sampled using the following distribution ${\hat{p}_{\text{land}}}^x \sim \text{U}\left(0.0,1.0\right)$m while the lateral desired distance is sampled uniformly with ${\hat{p}_{\text{land}}}^y \sim \text{U}\left(-0.3,0.3\right)$m.

The reward function proposed before is modified to allow jumps of different distances. In particular, the imitation reward (Eq. \eqref{eq:imitation-reward}) is modified by setting the weight for the base position to zero ($w^p$), thus only the joint angles and the orientation imitation are rewarded. The penalties for landing velocity and smoothing, as well as the survival reward, are kept (Eq. \eqref{eq:smoothing-pen}$\sim$\eqref{eq:reward-survival}). Similar to the imitation stage, a reward is given for landing close to the target (Eq. \eqref{eq:landing-reward}), except that the desired landing position $\boldsymbol{\hat{p}}_{\text{land}}$ is a varying input to the neural network. 

An additional reward is given for matching the desired base velocity in the $x-y$ direction, which is approximated by $\boldsymbol{\hat{v}}_{\text{xy}} = \boldsymbol{\hat{p}}_{\text{land}}/T$ with $T$ being the desired jumping duration that is extracted from the demonstration. Once landed, the desired velocity is set to be 0 m/s. The desired velocity is tracked by
\begin{equation}
    r^{vel}_i = w^v f(k_{v},\boldsymbol{\hat{v}}_{\text{xy}},\boldsymbol{v}_{\text{xy}}).
\end{equation}
with $w^v$ being the weight.

It should be mentioned that the termination conditions have also been revised at this stage. In contrast to the imitation stage, the episode is no longer terminated if the contact state does not match that of the demo, considering that a different jumping distance may require a varying stance and flight phase durations. Alternatively, an episode is terminated if the landing error is above a threshold. To achieve better tracking performance and save training time, the threshold is a function of the desired jumping distance. As a result, the early termination will be triggered when
\begin{equation}
    \left|\boldsymbol{\hat{p}}_{\text{land}}  - \boldsymbol{p}_{\text{land}} \right| > \alpha\left|\boldsymbol{\hat{p}}_{\text{land}}\right| + \beta,
    \label{eq:landing-termination-condition}
\end{equation}
where $\alpha$ and $\beta$ determine the allowable landing error. 


\subsection{Stage 3: Robust jumping stage}
\label{Sec: Uneven Terrain Stage}
Building upon the policies trained during the generalization phase, this stage aims for robust jumping over uneven terrains.
Differing from the work in \cite{bellegarda2024robust} and \cite{li2023learning}, the uneven terrain setup in our simulation comprises boxes of varying heights, bringing in a challenge to the robot's stability. In particular, height perturbations are applied between the robot's front and rear legs during the initial state to prevent lateral falls. At the beginning of each episode, the joint angles are computed by inverse kinematics (IK), allowing a stable standing on the boxes with height variation. However, the robot can still experience height variation between the left and right legs during landing. Similar to the generalization stage, the robot is prompted to jump different distances, both in the forward and lateral direction.

The terrain is instantiated using boxes with a height of 15cm. At the beginning of each episode, a height perturbation is sampled from a uniform distribution $\text{U}(-\epsilon,+\epsilon)$ m and is added to the constant box height to emulate an uneven terrain. Furthermore, during training, there is a 20\% probability that no height perturbations are sampled, avoiding over-fitting the policy for jumping across uneven terrains. 

\begin{figure*}[h]
    \centering
    \begin{subfigure}[h]{1\textwidth}
        \includegraphics[width=\textwidth]{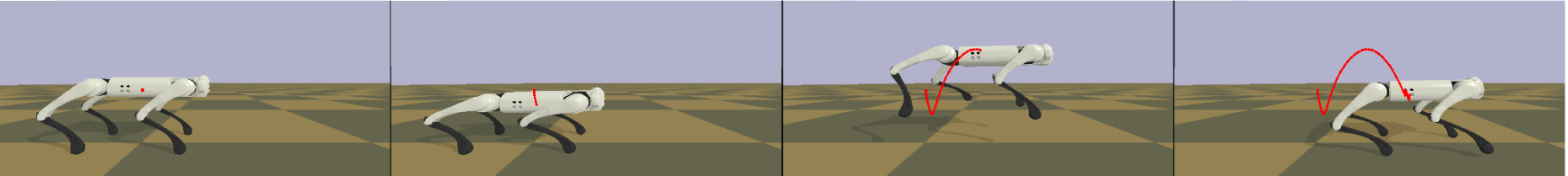}
    \end{subfigure}
    \begin{subfigure}[h]{1\textwidth}
        \includegraphics[width=\textwidth]{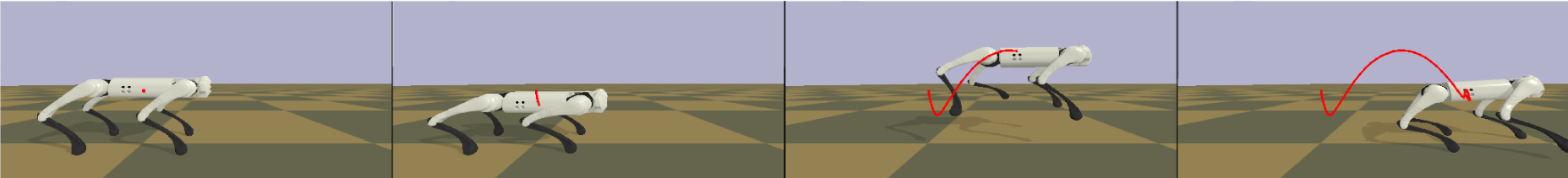}
    \end{subfigure}
        \begin{subfigure}[h]{1\textwidth}
        \includegraphics[width=\textwidth]{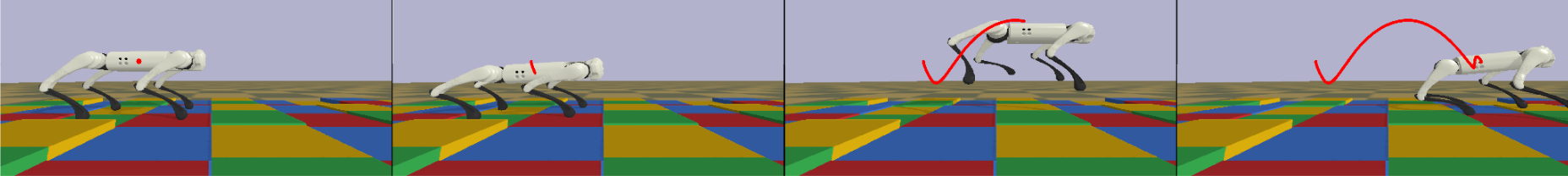}
    \end{subfigure}
    \caption{Learned quadrupedal jumping: (top) 35cm forward jumping on flat ground, (middle) 70cm forward jumping on flat ground, (bottom) 75cm over uneven terrain. The red curves plot the real CoM trajectories.\vspace{-1em}}
    \label{versatile_motion}
\end{figure*}

The reward defined for flat ground is also used for uneven terrain. This includes all the weights and length scales of the reward function as well as the termination limit parameters that were specified in the previous stage. The hyper-parameters for network training are also kept constant.

\subsection{Stage 4: Domain randomization stage}
\label{Sec: Domain randomization}

Domain randomization (DR) \cite{tan_sim--real_2018} helps to bridge the sim2real gap. Through varying parameters within both the environment and the robotic system itself, DR enables us to learn robust policy against uncertainties.

In this work, modelling parameters are randomized to make the model of the agent more realistic. To this end, the mass of the agent and its weight distribution are randomized. Furthermore, the center of mass (CoM) is perturbed by attaching an offset mass on the robot's base with different locations. To accommodate for the state error, we also randomize the initial state of the agent. Contact dynamics are also diversified by randomly sampling friction coefficients for the feet and ground, along with their corresponding restitution coefficients. Finally, the motor strength (a scaling factor applied on the actuation torque) and a frictional torque, are randomized to make the actuation model more realistic. For the system which utilizes elastic actuation, the spring parameters, including spring stiffness, spring damping, and rest angles, are randomized. 

\section{Results} \label{sec:result_discussion}
The section validates the learned jumping policy. In particular, we highlight the benefits of parallel compliance via comparison studies on jumping behavior and energy performance. All the results can be seen from: \textcolor{blue}{\url{https://youtu.be/FC_vLAc-lp0}}. 

\subsection{Training setup}
In this work, we use the Unitree Go1 robot to execute all the tasks. The training environment was built with PyBullet simulator\cite{coumans2021}, upon \cite{git_env}. In addition, this environment also emulates parallel springs attached to sagittal joints. The spring constants used for each joint have been chosen to match our design in \cite{ding2024quadrupedal} with $K_{\text{Thigh}} = 16$ and $K_{\text{Calf}} = 10$. 

We train the network using the proximal policy optimization (PPO) algorithm \cite{schulman2017proximal}, with both the actor and the critic parameterized by a two layer multi-layer perceptron (MLP) network with 1024 neurons at each layer and tanh activation functions between. Detailed PPO parameters are listed in Table~\ref{tab: PPO hyperparameters}.
\begin{table}
\caption{PPO parameters}
\centering
\resizebox{0.9\columnwidth}{!}{%
\begin{tabular}{|l|c|}
\hline
\textbf{RL Algorithm Parameters}                      & \textbf{Value} \\ \hline
Total Time-steps                                      & $10\times10^{6}$      \\ \hline
Linearly Decreasing Starting Learning Rate            & $1\times 10^{-4}$     \\ \hline
Batch Size                                            & 4096                  \\ \hline
Mini Batch Size                                       & 128                   \\ \hline
General Advantage Estimator (GAE) Lambda              & 0.95                  \\ \hline
Discount Factor $(\gamma)$                            & 0.99                  \\ \hline
Number of Epochs                                      & 10                    \\ \hline
Clip Range                                            & 0.2                   \\ \hline
Maximum Gradient Norm                                 & 0.5                   \\ \hline
Value Function Coefficient                            & 0.5                   \\ \hline
Neurons per Layer                            & (1024, 1024)                      \\ \hline
Activation Function                          & tanh                           \\ \hline
\end{tabular}%
}
\label{tab: PPO hyperparameters}
\vspace{-6mm}
\end{table}

\begin{figure*}[t]
    \centering
    \begin{minipage}{0.5\textwidth}
    \centering
    \includegraphics[width=0.95\linewidth,draft=False]{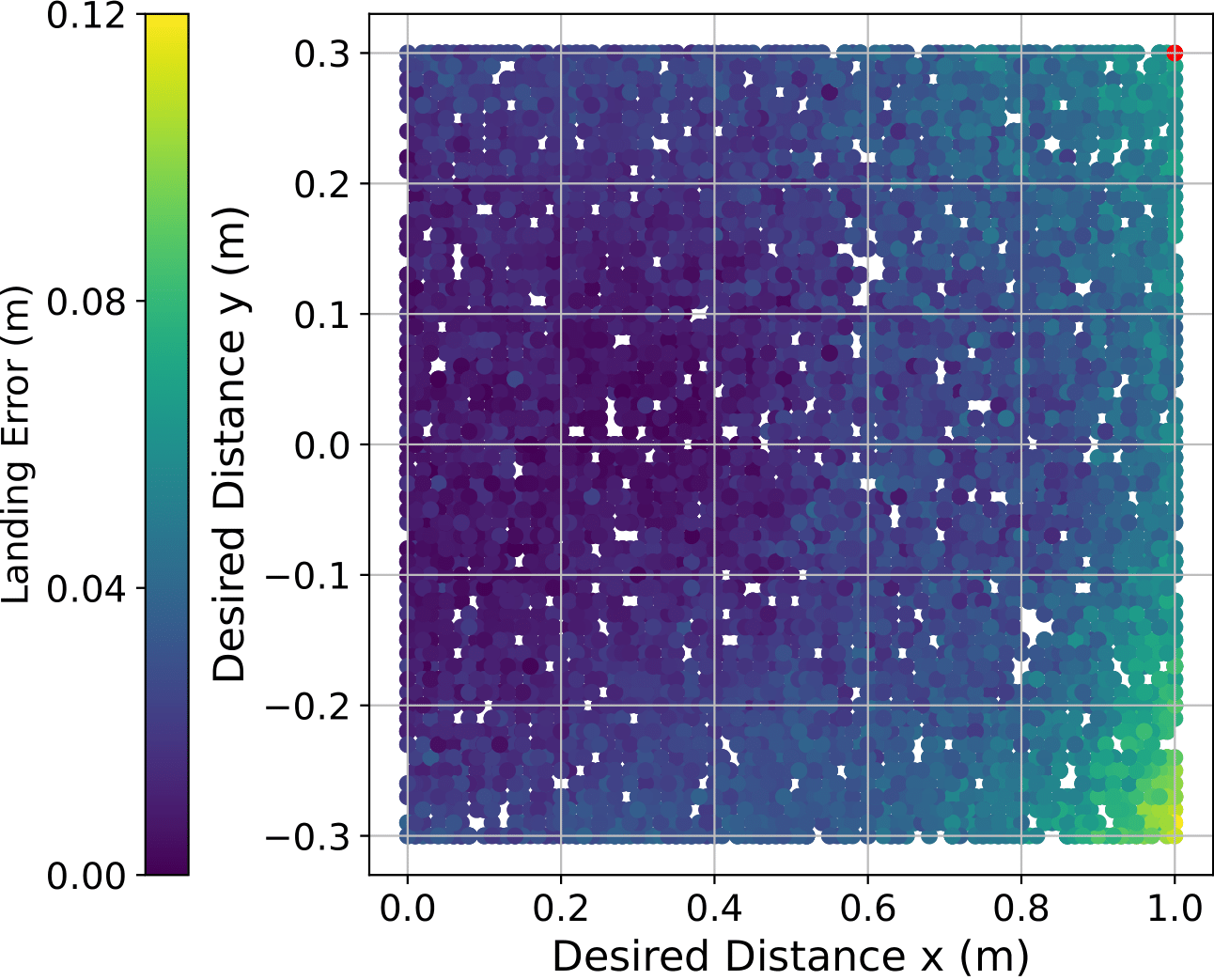}
    \label{fig:no-pea-landing-error-vs-desired-pos}
    \end{minipage}%
    \hfill
    \begin{minipage}{0.5\textwidth}
    \centering
    \includegraphics[width=0.95\linewidth,draft=False]{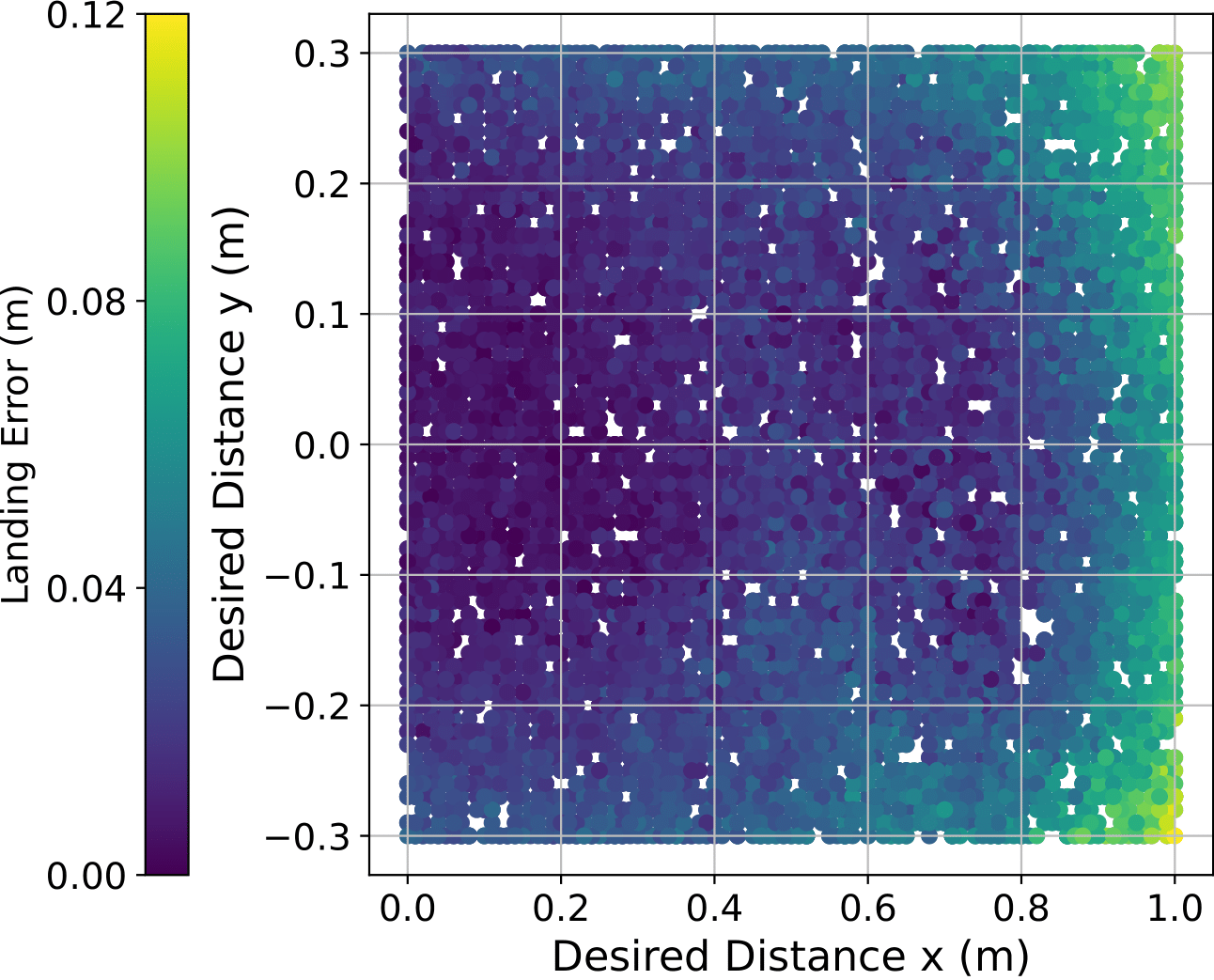}
    \label{fig:pea-landing-error-vs-desired-pos}
    \end{minipage}
    \caption{Landing errors for both rigid and compliant robots when jumping at different positions. The left side plots the results with the rigid robot while the right side plots the results with the compliant robot. The red dot marks the failed motion.}
    \label{fig:landing-error-vs-desired-jumping-distance}
    \vspace{-7mm}
\end{figure*}


\subsection{Versatile jumping}
We start by mimicking the demonstration for 40cm forward jump. By the end of this stage, both the rigid and soft robots can reproduce the demonstrated motion.  Then, in the 2$^{\text{nd}}$ stage, we generalized the learned policy. As a result, both agents can execute versatile jumps with varying landing positions, despite trained only with one single jump demonstration. Three motions are visualized in Fig.~\ref{versatile_motion}.  

\subsubsection{Tracking performance}  Fig.~\ref{fig:landing-error-vs-desired-jumping-distance} plots the landing errors for 10000 trails for both systems. We can find that, as the forward jumping distance (marked by the `Desired distance x') increases, the landing error rises for both systems. However, the symmetric lateral jumping is observed and the variation of lateral distance (marked by the `Desired distance y') does not affect the landing error a lot. Notably, unsuccessful jumps only occur in the rigid case.

\begin{figure}
    \centering
    \includegraphics[width=0.95\columnwidth]{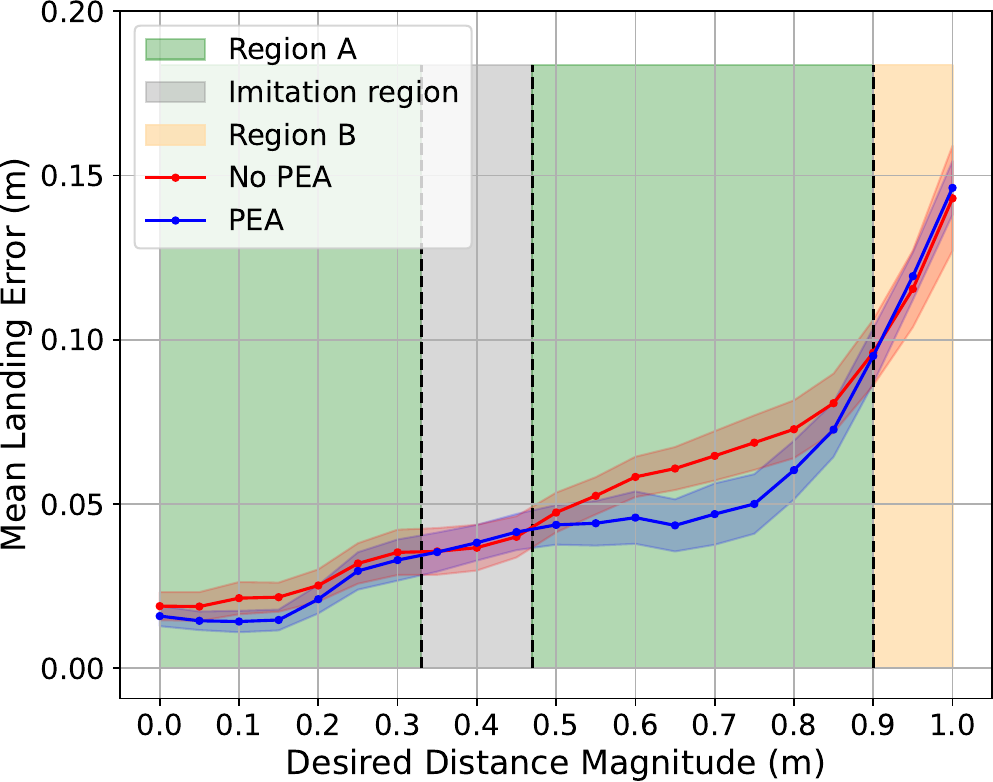}
    \caption{Landing error across jumping distances. Solid lines with dots indicate the mean, while shaded regions represent the 95\% confidence interval. 'PEA' and 'No PEA' correspond to the compliant and rigid systems, respectively.}
    \label{fig:desired-distance-magnitude-vs-mean-landing-error}
\end{figure}

 Fig.~\ref{fig:desired-distance-magnitude-vs-mean-landing-error} shows the mean landing error and its 95\% confidence interval with respect to the desired landing distance norm, in intervals of 5 cm. We can see that the mean landing errors are almost identical for both rigid and compliant systems when the desired landing position is close to 0.4m (see the `imitation region' in grey) or when the landing position is above 0.9m (`Region B' in yellow). However, in the green region (`Region A') that covers most distance, the compliant system has a marginally lower mean landing error. The average error across the entire jumping range is 0.054m for the rigid robot and 0.048m for the soft robot, reflecting an 11.1\% improvement by exploiting parallel compliance.
\begin{figure}
    \vspace{-2mm}
    \centering
    \includegraphics[width=0.93\columnwidth]{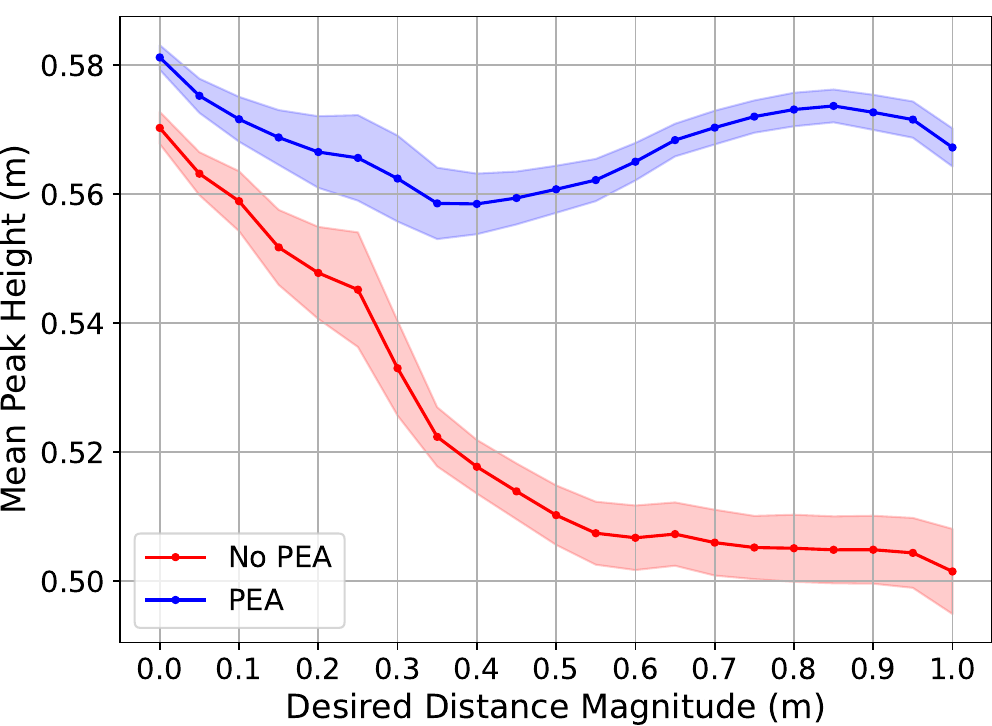}
    \caption{Variation of peak height over distance for both systems. The shaded region indicates the 95\% confidence interval. 
    }
    \label{fig:peak-height-vs-desired-jumping-distance-magnitude}
\end{figure}

\subsubsection{Peak height} 
The peak height, which is another key character for explosive jumping, is visualized in Fig.~\ref{fig:peak-height-vs-desired-jumping-distance-magnitude}.
As can be seen, parallel compliance contributes to a higher peak height due to the release of the potential energy that is stored when the robot squats down in the stance phase. The statistical result of the 10000 jumps demonstrates that the system with PEA jumps on average 8.8\% higher than the one without PEA. Besides, Fig.~\ref{fig:peak-height-vs-desired-jumping-distance-magnitude} reveals that the rigid robot decreases its peak height for a longer jump while the robot with parallel elasticity preserves its peak height.

\begin{figure}
    \centering
    \includegraphics[width=.9\columnwidth]{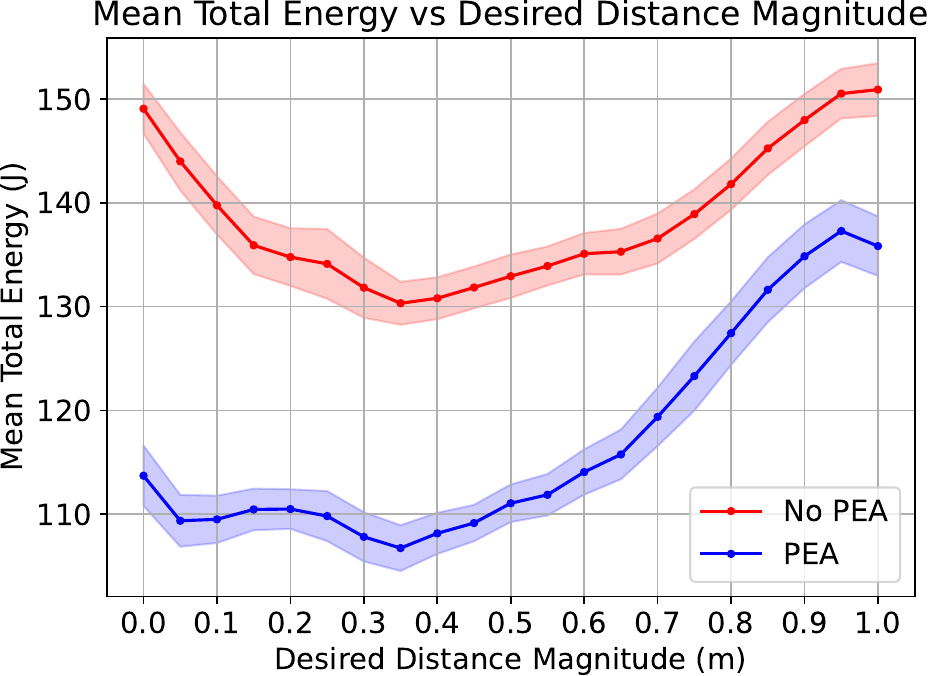}
    \caption{Illustrates how total energy varies against the desired
distance magnitude for the systems with and without PEA. The shaded region indicates the 95\% confidence interval.}
    \label{fig:energy-vs-desired-distance-magnitude}
    \vspace{-4mm}
\end{figure}

\begin{figure}
    \centering
    \includegraphics[width=0.95\columnwidth]{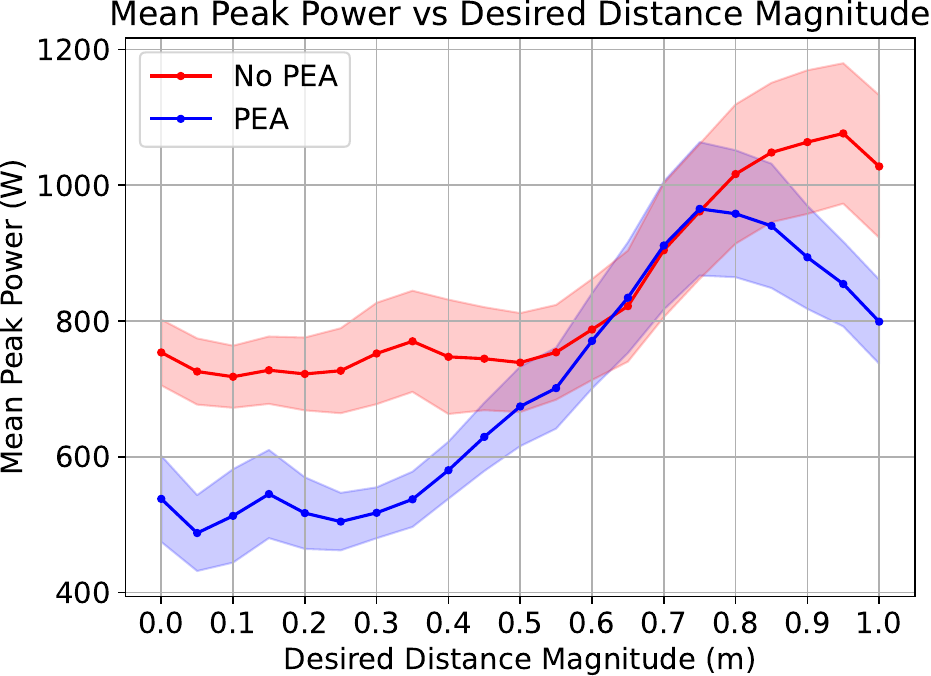}
    \caption{Peak power variation with the jumping distance.\vspace{-1em}}
    \label{fig:peak-power-vs-desired-distance-magnitude}
\end{figure}
\subsubsection{Total energy \& peak power} 
Fig.~\ref{fig:energy-vs-desired-distance-magnitude} and Fig.~\ref{fig:peak-power-vs-desired-distance-magnitude} separately plot the total energy consumption and peak power\footnote{We here compute the mechanical energy cost and mechanical power.} against the varying jumping distance. 

From Fig.~\ref{fig:energy-vs-desired-distance-magnitude}, we can see that the energy cost drops when the distance increases from 0m to 0.35m and increases when the distance increases from 0.4m to 1m. No matter with which jumping distance, the energy consumption for the system without parallel elasticity is higher than the system with PEA.
The average energy cost for 10000 jumps from 0 to 1m is 137.2J with stiff actuation while 116.4J with PEA, meaning a reduction of 15.2\%. 

Fig.~\ref{fig:peak-power-vs-desired-distance-magnitude}) shows that for jumps less than 0.6m the compliant robot requires less peak power. For jumps between 0.6$\sim$0.75m the compliant and rigid robots experience approximately the same peak power. For a jump longer than 0.75m, the peak power for the compliant system with PEA decreases as the jumping distance increases whereas the stiff system still experiences higher peak powers. The statistical results on 10000 jumps reveal that the rigid robot without parallel compliance experiences a mean peak power of 833.8W. In comparison, the system with PEA experiences a mean peak power of 702.1W, indicating a 15.8\% decrease.


\begin{figure*}[t]
    \centering
    \begin{minipage}{0.5\textwidth}
    \centering
    \includegraphics[width=0.9\linewidth,draft=False]{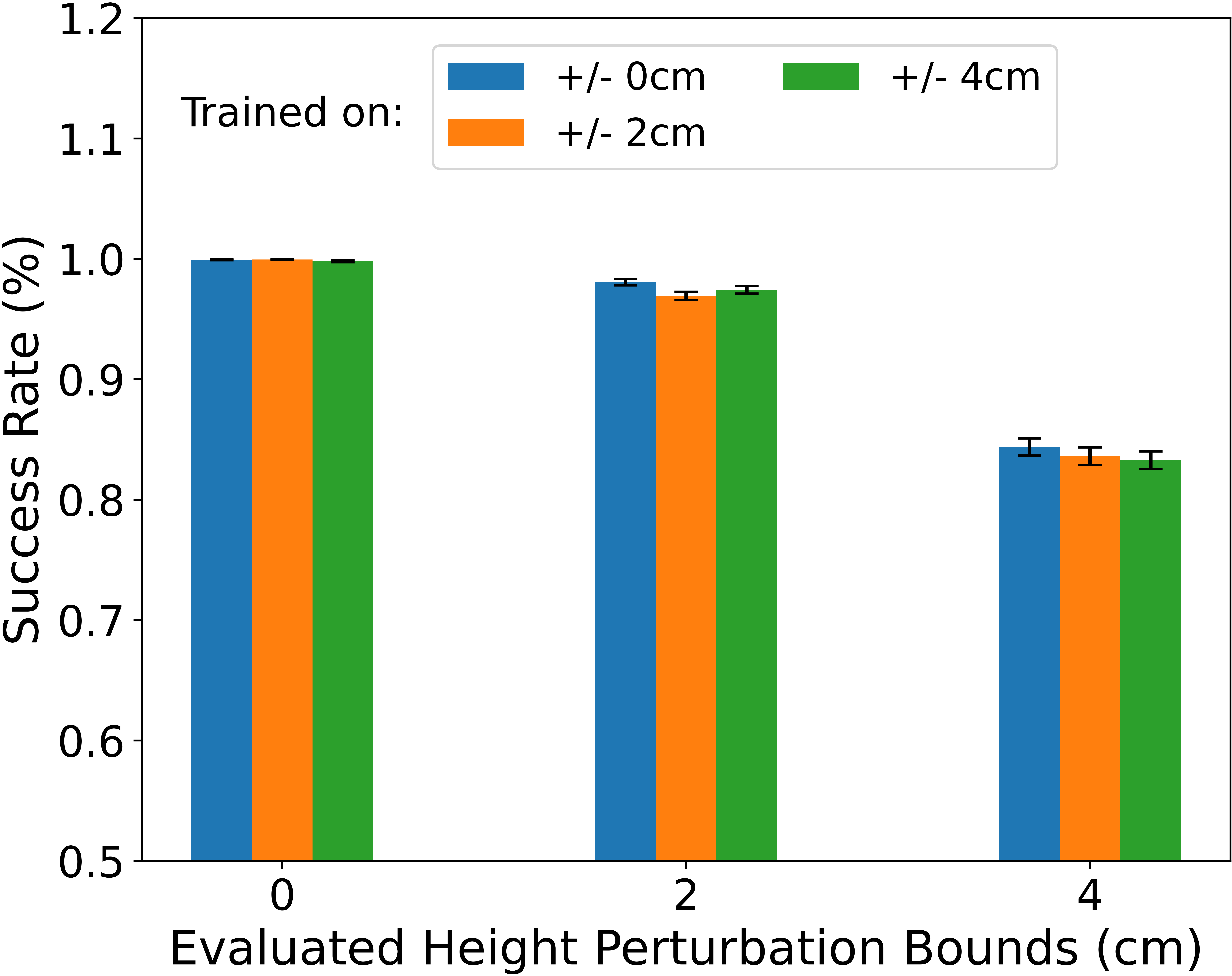}
    \label{fig:uneven_terrain_success_rate}
    \end{minipage}%
    \begin{minipage}{0.5\textwidth}
    \centering
    \includegraphics[width=0.9\linewidth,draft=False]{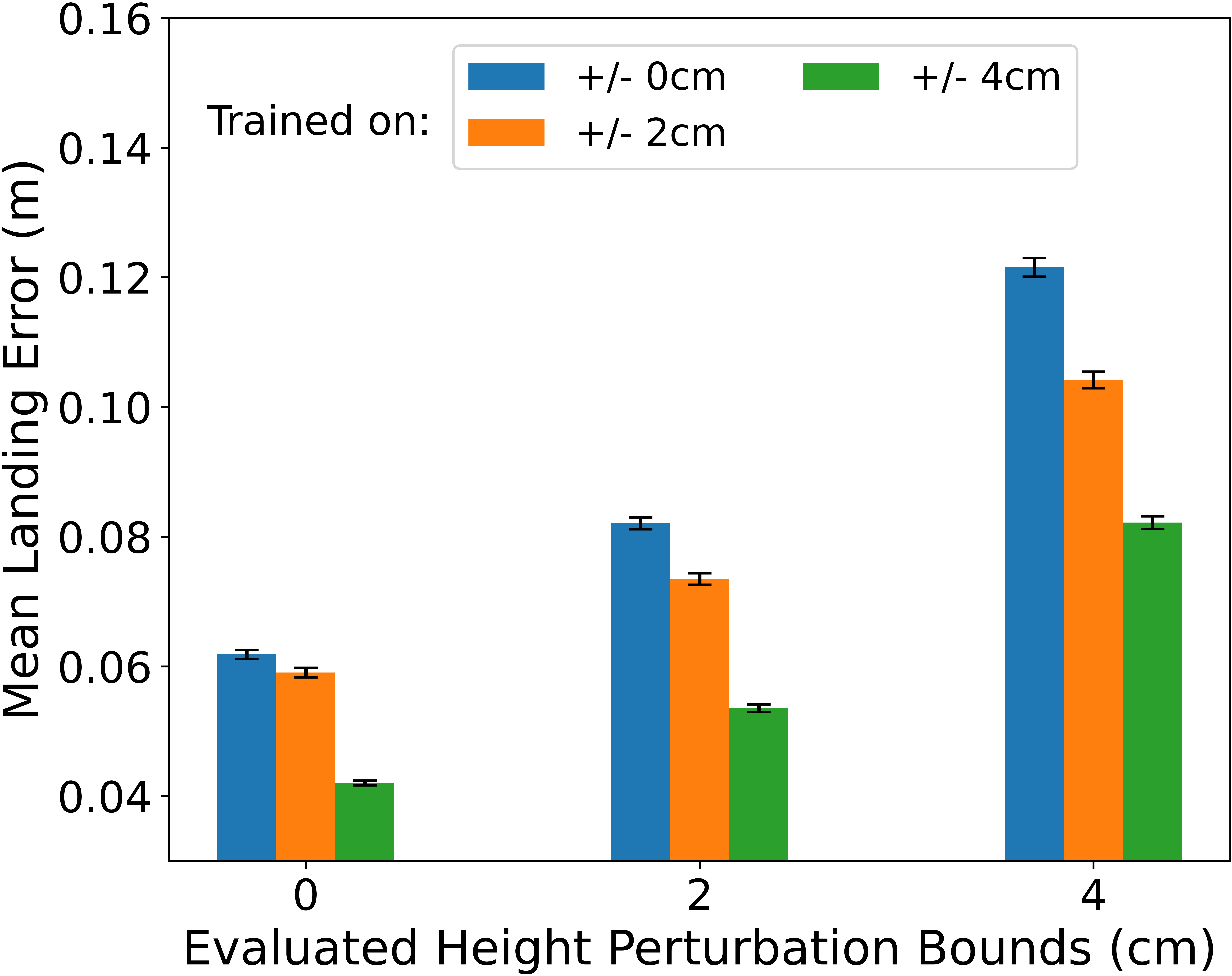}
    \label{fig:uneven_terrain_landing_error}
    \end{minipage}
    \caption{The comparison of success rate (left) and landing error (right) for rigid jumping with three policies: 1) a flat ground policy, 2) a policy trained over $\pm$2cm uneven terrain and 3) a policy trained over $\pm$4cm uneven terrain.\vspace{-1em}}
    \label{fig:uneven-terrain-results}
\end{figure*}

\begin{figure*}[h]
    \centering
    \begin{minipage}{0.5\textwidth}
    \centering
    \includegraphics[width=0.9\linewidth,draft=False]{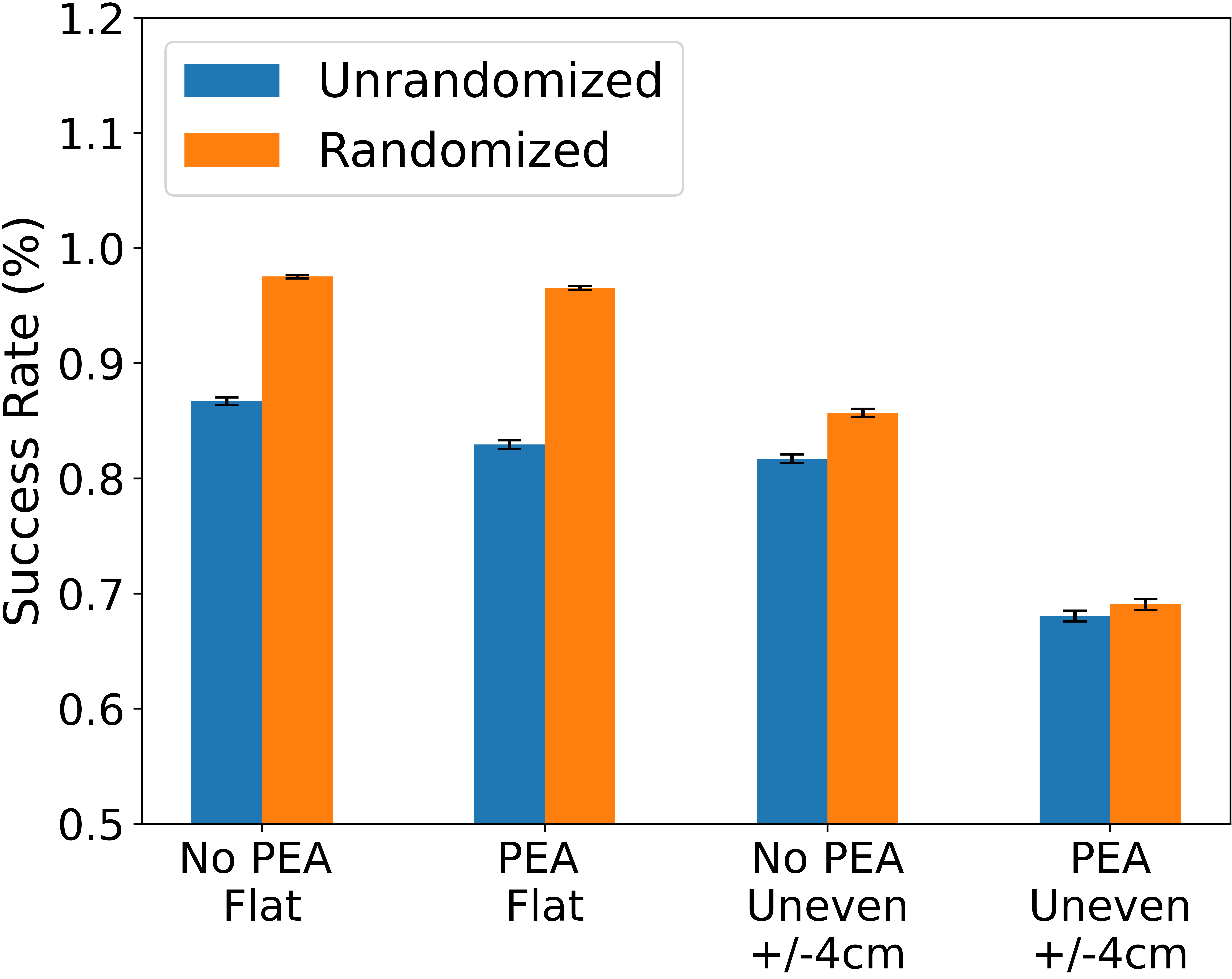}
    \label{fig:uneven_terrain_success_ratex}
    \end{minipage}%
    \begin{minipage}{0.5\textwidth}
    \centering
    \includegraphics[width=0.9\linewidth,draft=False]{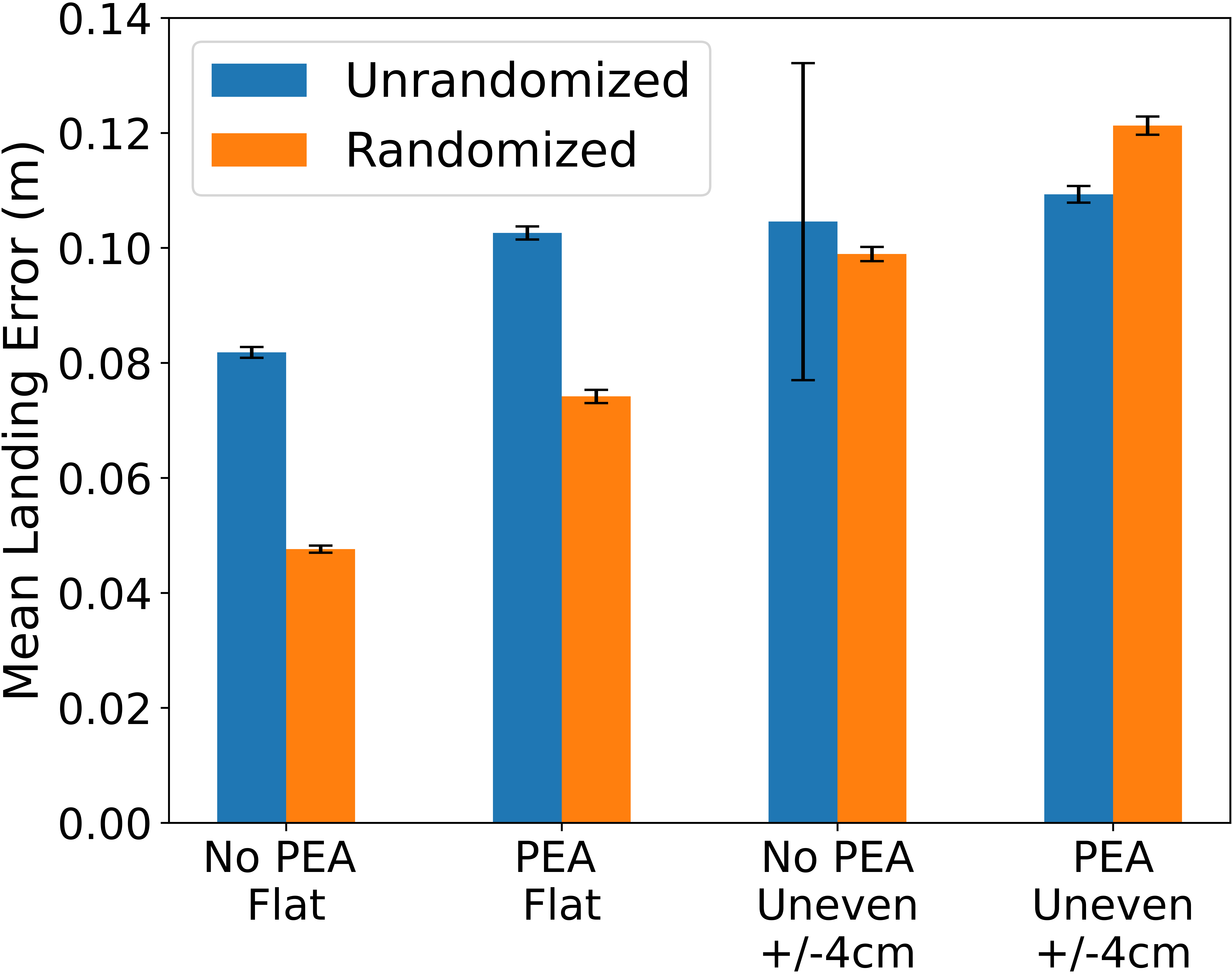}
    \label{fig:uneven_terrain_landing_errorx}
    \end{minipage}
    \caption{Comparison of (left) success rate and (right) landing error in a randomized environment with the policies before and after domain randomization.
    \vspace{-1em}}
    \label{fig:domain-randomisation-results}
\end{figure*}

\subsection{Robust jumping}
Within the 3$^{\text{th}}$ and 4$^{\text{th}}$ stages, the agent is trained for jumping over uneven terrains. The difficulty of one scenario is dictated by the maximum height variations ($\epsilon$) added to the terrain. 

\subsubsection{Jumping across uneven terrain}
Taking the rigid robot as an example, Fig.~\ref{fig:uneven-terrain-results} compares three policies: 1) the flat terrain policy trained with flat ground, 2) uneven terrain policy trained with  $\epsilon$ = 2cm, and 3) uneven terrain policy  $\epsilon$ = 4cm, in terms of the success rate and landing error. As visualized in Fig.~\ref{fig:uneven-terrain-results} (left), the success rate decreases for all three policies as the height perturbation increases. Fig.~\ref{fig:uneven-terrain-results} (right) reveals that, as the difficulty of the terrain increases, the landing error also increases. This implies that as the environment becomes more difficult the jumps become less precise and prone to fail. Further observation reveals that, by training a policy over the high-level terrain, the landing error in one certain evaluation environment decreases while the success rate does not change a lot. Among the three policies, the policy trained on perturbations of $\pm$4cm achieves the smallest landing error across all evaluated environments. 

\subsubsection{Domain randomization} 
Domain randomization was performed on four cases, including the rigid and compliant jumping over uneven terrains ($\epsilon$ = 4cm) as well as the rigid and compliant jumping on flat ground. A comparison between the baseline policies (`Unrandomized') and their randomized versions is depicted in Fig.~\ref{fig:domain-randomisation-results}. 

 As can be seen from Fig.~\ref{fig:domain-randomisation-results}, the randomized policies have a larger success rate than the ones without randomization. Also, the smaller landing error is also achieved, except compliant jumping on uneven terrain. The policy with the highest variation in landing error is the policy over uneven terrain before domain randomization while the policy with the lowest landing error is the randomized flat terrain policy without PEA.

\section{Conclusion}\label{sec:conclusion}
To conclude, this paper introduces an example-guided RL scheme for explosive jumping with both rigid and articulated compliant robots. The proposed scheme starts by mimicking a coarse reference trajectory generated by trajectory optimization, using a reduce-order dynamical model. By merely learning from a single demonstration, we realize versatile jumping with varying landing distances by conditioning the trained policy to a specific target. Furthermore, robust jumping over uneven terrain is realized by exposing the agent to uneven terrain during the training process.

Our work also highlights the benefits of parallel compliance. Results demonstrate that, with parallel springs, the robot reduces the mean energy cost by 15.2\% and drops the peak power by 15.8\%. The peak height of the explosive jumping increases by 8.8\% by exploiting parallel compliance and the mean landing error decreases by 11.1\%.

In the future, we will apply the learned policy to real robots.

\bibliographystyle{IEEEtran}
\bibliography{ref.bib}

\end{document}